\def\year{2020}\relax
\documentclass[letterpaper]{article} 
\usepackage{aaai20}  
\usepackage{times}  
\usepackage{helvet} 
\usepackage{amsfonts}
\usepackage{courier}  
\usepackage[hyphens]{url}  
\usepackage{graphicx} 
\usepackage{graphicx}  
\title{Accelerating Column Generation via Flexible Dual Optimal Inequalities with Application to Entity Resolution}

\usepackage{array}
\usepackage{booktabs}
\usepackage{multirow}
\usepackage{comment}
\usepackage{graphicx}
\usepackage{amsmath}
\usepackage[utf8]{inputenc}
\usepackage[english]{babel}
\usepackage{algorithm}
\usepackage{algpseudocode}
\usepackage{amsthm}
\usepackage{enumitem}
\usepackage{bm}
\usepackage{url}

\newcommand{\citet}[1]{\citeauthor{#1} \shortcite{#1}}
\newcommand{\citep}{\cite}

\author{Vishnu Suresh Lokhande\textsuperscript{\rm 1,\rm 3} \quad  Shaofei Wang\textsuperscript{\rm 2} \quad Maneesh Singh\textsuperscript{\rm 3} \quad
	Julian Yarkony\textsuperscript{\rm 3}\\
	\hspace{0.5in} \small\textrm{lokhande@cs.wisc.edu} \hspace{0.4in} \small\textrm{sfwang@seas.upenn.edu} \hspace{0.1in} \small\textrm{msingh@verisk.com} \hspace{0.1in} \small\textrm{julian.e.yarkony@gmail.com}\\\\
\textsuperscript{\rm 1}University of Wisconsin-Madison\\ 
\textsuperscript{\rm 2}University of Pennsylvania\\ 
\textsuperscript{\rm 3}Verisk Computational and Human Intelligence Laboratory\\ 
}

\newcommand\ie{\emph{i.e.}}

\begin{document}

\maketitle

\begin{abstract}
In this paper, we introduce a new optimization approach to Entity Resolution. Traditional approaches tackle entity resolution with hierarchical clustering, which does not benefit from a formal optimization formulation. In contrast, we model entity resolution as correlation-clustering, which we treat as a weighted set-packing problem and write as an integer linear program (ILP). 
In this case, sources in the input data correspond to elements and entities in output data correspond to sets/clusters.  We tackle optimization of weighted set packing by relaxing integrality in our ILP formulation. 
The set of potential sets/clusters can not be explicitly enumerated, thus motivating optimization via column generation.  In addition to the novel formulation, we also introduce new dual optimal inequalities (DOI), that we call flexible dual optimal inequalities, which tightly lower-bound dual variables during optimization and accelerate column generation.  We apply our formulation to entity resolution (also called de-duplication of records), and achieve state-of-the-art accuracy on two popular benchmark datasets. Our F-DOI can be extended to other weighted set-packing problems as well. The project page is available at the following url,\\ \textit{https://github.com/lokhande-vishnu/EntityResolution}
\end{abstract}

\section{Introduction}
\label{sec:intro}

 In this paper we study the problem of entity resolution. Entity resolution aims to eliminate redundant information from multiple sources of data. This task plays a key role in information integration, natural language understanding, information processing on the World-Wide Web all of which are core areas of AI \citep{konda2016magellan}.

 \subsubsection{The Problem Setup} Given a dataset of observations each associated with up to one object, entity resolution aims to pack (or partition) the observations into groups called hypothesis (or entities) such that there is a bijection from hypotheses to unique entities in the dataset.  We are provided a set of observations called records, where each record is associated with a subset of fields (for example:  name, social security number, phone number etc).  We seek to partition the observations into hypothesis so that: \textbf{(1)} all observations of any real world entity are associated with exactly one selected hypothesis; \textbf{(2)} each selected hypothesis is associated with observations of exactly one real world entity.  
 
 \begin{figure}[!t]
\centering
\includegraphics[width=0.75\columnwidth]{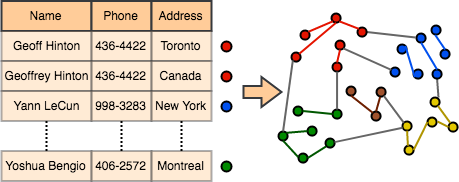} 
\caption{\textbf{Entity Resolution posed as MWSP:} Given a structured (tabular) data, we study the problem of entity resolution where we group rows representing the same real-world entity into the same cluster. Entity Resolution can be viewed as a node-clustering problem over a graph. In the example used in this figure, we identify similar rows with identical colored dots and then tranform it into a graph with nodes representing the entities/rows and edges represented some measure of similarity between between the nodes. Nodes that are not connected are understood to represent distict rows in the table.} 
\label{fig:ERasMWSP}
\end{figure}

 \subsubsection{Traditional Approaches } Entity resolution has been studied using different clustering approaches \citep{saeedi2017comparative}. It is common to transform entity resolution to a graph problem and run a clustering algorithms on top of it as depicted in Figure~\ref{fig:ERasMWSP}. The popular clustering algorithms developed to attack entity resolution are ConCom, where the algorithm is based on computing the connected components of the input graph.  Center clustering sequentially adds edges from a priority queue and either assigns the nodes to a cluster or tags them as a center \citep{hassanzadeh2009creating}.  Star clustering \citep{aslam2004star}, in a similar way, prioritizes in adding those nodes to a cluster that have the highest degree.  Correlation Clustering \citep{corclustorig}, which forms the backbone of our method, has also been studied for entity resolution problem. However, the lengthy and numerous iterations to converge made it difficult for entity resolution problems \citep{saeedi2017comparative}. 
 
%
%
 %
 
\subsubsection{Entity Resolution as MWSP} Contrary to previous works, we propose to tackle entity resolution as an optimization problem, formulating it as a minimum weight set packing (MWSP) problem.  The set of all possible hypotheses is the power set of the set of the observations.  The real valued cost of a hypothesis, is a second order function of the observations that compose the hypothesis.  
The cost of a hypothesis decreases as the similarity among the observations in the hypothesis increases. Any non-overlapping subset of all possible hypotheses corresponds to a partition; we treat each observation not in any element in the subset as being in a hypothesis by itself.  We model the quality of a packing as the total cost of the hypothesis in the packing.  The lowest total cost packing is empirically a good approximation to the ground truth.

\subsubsection{Efficient MWSP using Column Generation} Enumerating the power set of the observations is often not possible in practice, thus motivating us to tackle MWSP using column generation (CG) \citep{cuttingstock,barnprice,desrosiers2005primer,lubbecke2010column,mwspJournal}.  CG solves a linear programming (LP) relaxation of MWSP by constructing a small sufficient subset of the power set, such that solving the LP relaxation over the sufficient subset provably provides the same solution as solving the LP relaxation over the entire power set.  CG can often be accelerated using dual optimal inequalities (DOIs) \citep{ben2006dual}, which bound the otherwise unbounded dual variables of the LP-relaxation, drastically reducing the search space of the LP problem.  The use of DOI provably does not alter the solution produced at termination of CG.  

\subsubsection{Core Contribution} We make the following contributions to the scientific literature.  
   \textbf{(1). }Introduce a novel MWSP formulation for entity resolution, that achieves efficient exact/approximate optimization using CG.  
    \textbf{(2). } Introduce novel DOIs called Flexible DOIs (F-DOI), which can be applied to broad classes of MWSP problems.  
 
\subsubsection{Paper Organization}  The paper is structured as follows.  In Section \ref{SEC_relWork} we review the integer linear programming (ILP) formulation of MWSP, and its solution via CG.  In Section \ref{SEC_stab_mwsp} we introduce F-DOIs.  In Section \ref{SEC_Applications} we devise optimization algorithms to solve entity resolution problem via CG and F-DOIs.  
 In Section \ref{SEC_Exper} we demonstrate the effectiveness of our approach on benchmark entity resolution datasets.  In Section \ref{SEC_conc} we conclude.
%
%
%
 %
%
%
\section{Preliminaries}
\label{SEC_relWork}
In this section we review the MWSP formulation and CG solution of \citep{mwspJournal}.  We outline this section as follows.  In Section \ref{Sec_review_ilp} we review the ILP formulation of MWSP.  In Section \ref{Sec_review_vanilla_cg} we review the CG algorithm that solves an LP relaxation of the ILP formulation.  In Section \ref{Sec_review_doi} we review the varying DOIs introduced in \citep{mwspJournal}.  To be consistent with the notation used in the operations research community, we use the notation of \citep{mwspJournal} throughout this paper. 
\subsection{ An ILP Formulation of MWSP }
\label{Sec_review_ilp}
\paragraph{Observations } An observation corresponds to an element in the traditional set-packing context 
and a data source in the  entity resolution context.
We use $\mathcal{D}$ to denote the set of observations, which we index by $d$.
\paragraph{Hypotheses } A hypothesis corresponds to a set in the traditional set-packing context, 
and an entity in the entity resolution context.  Given a set of observations $\mathcal{D}$, the set of all hypotheses is the power set of $\mathcal{D}$, which we denote as $\mathcal{G}$ and index by $g$.

We describe  $\mathcal{G}$ using matrix $G \in \{0,1\}^{|\mathcal{D}| \times |\mathcal{G}|}$.  Here $G_{dg}=1$ if and only if hypothesis $g$ includes observation $d$, and otherwise $G_{dg}=0$.  A real valued cost $\Gamma_g$ is associated to each $g \in \mathcal{G}$, where $\Gamma_{g}$ is the cost of including $g$ in our packing.  The hypothesis $g$ containing no observations is defined to have cost $\Gamma_g=0$. $\Gamma_g$ is instantiated as a function of $G_{dg}$. For example, let $d_1 \in \mathcal{D}, d_2 \in \mathcal{D}$ and let $\theta_{d_1d_2} \in \mathbb{R}$ denote the cost of putting $d_1, d_2$ in a single hypothesis $g$, then we could write $\Gamma_g=\sum_{d_1, d_2}\theta_{d_1d_2}G_{d_1g}G_{d_2g}$. We will discuss more on the cost ($\Gamma_g$) formulation, in the light of entity resolution, in Section~\ref{App_clust_form}.

A packing is described using $\gamma \in \{0,1\}^{|\mathcal{G}|}$ where $\gamma_g=1$ indicates that hypothesis $g$ is included in the solution, and otherwise $\gamma_g=0$.  MWSP is written as an ILP below.
\begin{align}
    \label{ILPVER}
    \min_{ \gamma \in \{0,1 \}^{|\mathcal{G}|} } \quad & \sum_{g \in \mathcal{G}}\Gamma_g\gamma_g \\
    \text{s.t.} \quad & \sum_{g \in \mathcal{G}}G_{dg}\gamma_g \leq 1 \quad \forall d \in \mathcal{D} \nonumber
\end{align}
The constraints in Eq \ref{ILPVER} enforce that no observation is included in more than one selected hypothesis in the packing.  
\subsection{Solving MWSP via Column Generation}
\label{Sec_review_vanilla_cg}
\subsubsection{Column Generation Algorithm  }
Solving Eq \ref{ILPVER} is challenging for two key reasons:  \textbf{(1)} MWSP is NP-hard  \citep{karp}; \textbf{(2)} $\mathcal{G}$ is too large to be considered in optimization for our problems.  To tackle \textbf{(1)}, the integrality constraints on $\gamma$ are relaxed, resulting in an LP:  

%
%
\begin{align}
    \label{LPVer}
    \mbox{Eq } \ref{ILPVER} \geq 
    \min_{ \gamma \geq 0 } \quad & \sum_{g \in \mathcal{G}}\Gamma_g\gamma_g \\
    \text{s.t.} \quad & \sum_{g \in \mathcal{G}}G_{dg}\gamma_g \leq 1 \quad \forall d \in \mathcal{D} \nonumber
\end{align}
%
 \citep{mwspJournal} demonstrates that \textbf{(2)} can be circumvented by using column generation (CG).  Specifically, the CG algorithm constructs a small sufficient subset of $\mathcal{G}$, (which is denoted $\hat{\mathcal{G}}$ and initialized empty) s.t. an optimal solution to Eq \ref{LPVer} exists for which only hypothesis in $\hat{\mathcal{G}}$ are used.  Thus CG avoids explicitly enumerating $\mathcal{G}$, which grows exponentially in $|\mathcal{D}|$.  
The primal-dual optimization over $\hat{\mathcal{G}}$, which is referred to as the restricted master problem (RMP), is written as:
\begin{align}
    \label{RMP_PRIMAL}
    & \min_{\gamma \geq 0} && \sum_{g \in \hat{\mathcal{G}}}\Gamma_g\gamma_g \\
    & \text{s.t.} \quad && \sum_{g \in \hat{\mathcal{G}}} G_{dg}\gamma_g \leq 1 \quad \forall d \in \mathcal{D} \nonumber \\
    \label{RMP_DUAL}
   = \quad & \max_{\lambda \leq 0 } && \sum_{d \in \mathcal{D}}\lambda_d \\
   & \text{s.t.} \quad && \Gamma_g- \sum_{d \in \mathcal{D}}G_{dg}\lambda_d \geq 0 \quad \forall g \in \hat{\mathcal{G}} \nonumber
\end{align}
%
The CG algorithm is described in Alg~\ref{BasicColGenAlg}.  CG solves the MWSP problem by alternating between:
\textbf{(1)} solving the RMP in Eq \ref{RMP_DUAL}
given $\hat{\mathcal{G}}$  (Alg~\ref{BasicColGenAlg}, line~\ref{alg_1_rmp}) and \textbf{(2)} Adding hypothesis in $\mathcal{G}$ to $\hat{\mathcal{G}}$, that have negative reduced cost given dual variables $\lambda$ 
(Alg~\ref{BasicColGenAlg},line~\ref{alg_1_pricing}). The selection of the lowest reduced cost hypothesis in $\mathcal{G}$ is referred to as pricing, and is formally defined as:
\begin{align}
\label{pricing_gen}
    \min_{g \in \mathcal{G}}\Gamma_g-\sum_{d \in \mathcal{D}}\lambda_d G_{dg}
\end{align}
Solving Eq \ref{pricing_gen} is typically tackled using a specialized solver exploiting specific structural properties of the problem domain \citep{cuttingstock,wang2018accelerating,zhang2017efficient}. 
In many problem domains pricing algorithms return multiple negative reduced cost hypothesis in $\mathcal{G}$. 
In these cases some or all returned hypotheses with negative reduced cost are added to $\hat{\mathcal{G}}$. 
\begin{algorithm}[!t]
 \caption{MWSP via Column Generation}
\begin{algorithmic}[1] 
\State $\hat{\mathcal{G}} \leftarrow \emptyset$
\label{alg_1_init}
\Repeat
\label{alg_1_start}
\State $\gamma,\lambda \leftarrow $ Solve the RMP  in Eq \ref{RMP_PRIMAL}--\ref{RMP_DUAL} 
\label{alg_1_rmp}
\State $g^* \leftarrow$ Solve the pricing problem  in Eq \ref{pricing_gen}
\label{alg_1_pricing}
\State $\hat{\mathcal{G}} \leftarrow \hat{\mathcal{G}} \cup \{ g^*\} $
 \Until{ $\Gamma_{g^*}-\sum_{d \in \mathcal{D}}\lambda_dG_{dg^*} \geq 0$ }
\label{alg_1_converge}
 \State $\gamma \leftarrow$ Solve MWSP in Eq \eqref{ILPVER} over $\hat{\mathcal{G}}$ instead of $\mathcal{G}$
 \label{alg_1_term}
\State Return $\gamma$
\label{alg_1_return}
\end{algorithmic}
\label{BasicColGenAlg}
\end{algorithm}
    
\subsubsection{Convergence of Column Generation}
CG terminates when no negative reduced cost hypotheses remain in $\mathcal{G}$ (Alg~\ref{BasicColGenAlg},line~\ref{alg_1_converge}).  CG does not require that the lowest reduced cost hypothesis is identified during pricing to ensure that Eq \ref{LPVer} is solved exactly; instead, Eq \ref{LPVer} is solved exactly as long as a  $g \in \mathcal{G}$ with negative reduced cost is produced at each iteration of CG if one exists.  

 If Eq \ref{RMP_PRIMAL} produces a binary valued $\gamma$ at termination of CG (\ie\ the LP-relaxation is tight) then $\gamma$ is provably the optimal solution to Eq~\ref{ILPVER}. However if $\gamma$ is fractional at termination of CG, an approximate solution to Eq~\ref{ILPVER} can still be obtained by replacing $\mathcal{G}$ in Eq~\ref{ILPVER} with $\hat{\mathcal{G}}$ (Alg~\ref{BasicColGenAlg},line \ref{alg_1_term}).  
\citep{mwspJournal} shows that Eq \ref{LPVer} describes a tight relaxation in practice; 
We refer readers interested in tightening Eq \ref{LPVer} to \citep{mwspJournal}, which achieve this using subset-row inequalities \citep{jepsen2008subset}.     
\subsection{Dual Optimal Inequalities (DOIs)}
\label{Sec_review_doi}
The convergence of Alg~\ref{BasicColGenAlg} often can be accelerated by providing bounds on the dual variables in Eq~\ref{RMP_DUAL} without altering the final solution of Alg~\ref{BasicColGenAlg}, thus limiting the dual space that Alg~\ref{BasicColGenAlg} searches over. 
We define DOI with $\Xi_d$ which lower bounds dual variables in Eq~\ref{RMP_DUAL} as $-\Xi_d \leq \lambda_d,  \forall d \in \mathcal{D}$.  The primal RMP in Eq~\ref{RMP_PRIMAL} is thus augmented with new primal variables $\xi$, where primal variable $\xi_d$ corresponds to the dual constraint $-\Xi_d \leq \lambda_d$.
\begin{align}
    \label{RMP_DOI_primal}
    &\min_{\substack{\gamma \geq 0 \\ \xi \geq 0}} \quad && \sum_{g \in \hat{\mathcal{G}}} \Gamma_g\gamma_g +\sum_{d \in \mathcal{D}}\Xi_d \xi_d \\
    &\text{s.t.} \quad && -\xi_d+\sum_{g \in \hat{\mathcal{G}}} G_{dg}\gamma_g \leq 1 \nonumber \\
=   \label{RMP_DOI_dual}
    \quad & \max_{-\Xi_d \leq \lambda_d \leq 0} \quad && \sum_{d \in \mathcal{D}}\lambda_d \\
    &\text{s.t.} \quad && \Gamma_g- \sum_{d \in \mathcal{D}}G_{dg}\lambda_d \geq 0 \quad \forall g \in \hat{\mathcal{G}} \nonumber
\end{align}
\subsubsection{Varying DOIs of \citep{mwspJournal} }
In the applications of \citep{mwspJournal}, the authors observed that the removal of a small number of observations rarely causes a significant change to the cost of a hypothesis in $\hat{\mathcal{G}}$.  This fact motivates the following DOIs, which are called varying DOIs.  

Let $\bar{g}(g,\mathcal{D}_s)$ be the hypothesis consisting of $g$ with all observations in $\mathcal{D}_s \subseteq \mathcal{D}$ removed.  Formally, $G_{d\bar{g}(g,\mathcal{D}_s)}=G_{dg}[d \notin \mathcal{D}_s], \forall d \in \mathcal{D}$, where $[]$ is the binary indicator function.  
Let $\epsilon$ be a tiny positive number.  
Varying DOI are computed as:  
%
%
\begin{align}
    \label{better_journal_bound}
    \Xi_d = \epsilon+ \max_{\substack{g \in \hat{\mathcal{G}}}}\Xi^*_{dg} \quad \forall d \in \mathcal{D}\\
    \Xi^*_{dg} \geq \max_{\substack{\hat{g} \in \mathcal{G} \\ G_{\hat{d}\hat{g}}\leq G_{\hat{d}g} \forall \hat{d} \in \mathcal{D}}} \Gamma_{\bar{g}(\hat{g},\{d\})}-\Gamma_{\hat{g}}\nonumber
\end{align}
Observe that $\Xi_d$ may increase (but never decrease) over the course of CG as $\hat{\mathcal{G}}$ grows.  
In \citep{mwspJournal} the computation of $\Xi^*_{dg}$ is done using problem specific worst case analysis for each $g$ upon addition to $\hat{\mathcal{G}}$.  

%
%

\subsubsection{Related Approaches}
We now contrast DOIs from other dual stabilization methods, which also aim at accelerating CG. Dual stabilization approaches (excluding DOI) can all be understood as imposing a norm on the dual variables to prevent them from becoming extreme or leaving the area around a well established dual solution.  DOI based methods, in contrast, are based on providing provable bounds on the optimal point in the dual space.\\
\textit{\citep{du1999stabilized}}: This work optimizes the RMP with an $\ell_1$ penalty on the distance from a box around the best dual solution found thus far. Here best is defined as the maximum lower bound identified thus far over the course of column generation. Variants on this approach are available and provide different schedules for weakening the $\ell_1$ penalty. Other variants can replace the best solution found thus far with the most recent solution. The similar work of \citep{marsten1975boxstep} binds the dual variables to lie in a box around the previous dual solution.\\
\textit{\citep{gschwind2016dual}}: This work derives bounds on dual variables corresponding to swapping elements in hypotheses with other elements in the primal problem. It is appropriate for tasks such as bin packing and cutting stock where cost terms are not defined in terms of the elements that make up a set. In contrast, the varying DOI and F-DOI describe bounds on the dual variables corresponding to removing elements from hypotheses in the primal problem.  

\section{Flexible Dual Optimal Inequalities}
\label{SEC_stab_mwsp}
A major drawback of varying DOI 
is that $\Xi_d$ depends on all hypotheses in $\hat{\mathcal{G}}$ (as defined in Eq~\ref{better_journal_bound}), while often only a small subset of $\hat{\mathcal{G}}$ are active (selected) in an optimal solution to Eq~\ref{RMP_PRIMAL}.  
Thus during Alg~\ref{BasicColGenAlg}, the presence of a hypothesis in $\hat{\mathcal{G}}$ may increase the cost of the optimal solution found in current iteration, making exploration of solution space slower.  This motivates us to design new DOIs that circumvent this difficulty, which we name Flexible DOIs (F-DOIs). 

%
We outline this section as follows.  In Section \ref{FormMWSPNOvle} we introduce a MWSP formulation using CG featuring our F-DOIs.  In Section \ref{novelReduce} we consider pricing under this MWSP formulation.
\subsection{Formulation with F-DOIs}
\label{FormMWSPNOvle}
Given any $g \in \mathcal{G}$, let $\Xi_{dg}$ be positive if $G_{dg}=1$ and otherwise $\Xi_{dg}=0$, and defined such that for all non-empty $\mathcal{D}_s \subseteq \mathcal{D}$ the following bound is satisfied.
\begin{align}
\label{novelDoiBound}
   \sum_{d \in \mathcal{D}_s} \Xi_{dg}\geq \epsilon+\Gamma_{\bar{g}(g,\mathcal{D}_s)}-\Gamma_{g}
\end{align}
Let $\mathcal{Z}_d$ be the set of unique positive values of $\Xi_{dg}$ over all $g \in \hat{\mathcal{G}}$, which we index by $z$.  We order the values in $\mathcal{Z}_d$ from smallest to largest as $[\omega_{d1},\omega_{d2},\omega_{d3}...]$.    We describe $\Xi_{dg}$ using $Z_{dzg} \in \{0,1\}$ where $Z_{dzg}=1$ if and only if $\Xi_{dg}\geq \omega_{dz}$.  We describe $\Xi_{dg}$ using $\Xi_{dz}$ as follows:  
%
%
%
$\Xi_{dz}=\omega_{dz}-\omega_{d(z-1)} \quad \forall  z \in \mathcal{Z}_d,z\geq 2$;   
$\Xi_{d1}=\omega_{d1} $.   
%
Below we use $Z$ to model MWSP as a primal/dual LP. 
\begin{align}
\label{RMP_new_primal}
    & \min_{\substack{\gamma \geq 0  \\ \xi \geq 0}} && \sum_{g \in \hat{\mathcal{G}}}\Gamma_g\gamma_g +\sum_{\substack{d \in \mathcal{D} \\ z \in \mathcal{Z}_d }}\Xi_{dz} \xi_{dz} \\
    & \text{s.t.} && -\xi_{dz}+\sum_{g \in \hat{\mathcal{G}}}Z_{dzg}\gamma_g \leq 1 \quad \forall d \in \mathcal{D},z \in \mathcal{Z}_d \nonumber \\
    \label{RMP_new_dual}
=   &\max_{ \substack{-\Xi_{dz} \leq \lambda_{dz} \leq 0 \\ \forall d \in \mathcal{D}, z \in \mathcal{Z}_d
    } } && \sum_{\substack{d \in \mathcal{D} \\ z \in \mathcal{Z}_d }} \lambda_{dz} \\
    &\text{s.t.} &&\Gamma_g - \sum_{\substack{d \in \mathcal{D} \\ z \in \mathcal{Z}_d }} Z_{dzg}\lambda_{dz} \geq 0\quad \forall g \in \hat{\mathcal{G}} \nonumber
\end{align}
F-DOIs are the inequalities $-\Xi_{dz} \leq \lambda_{dz}$ in Eq \ref{RMP_new_dual}. 
%
We now prove that at termination of CG that $\xi_{dz}=0$ $\forall d\in \mathcal{D}, z \in \mathcal{Z}_d$ and hence Eq \ref{RMP_new_primal}=Eq \ref{LPVer}.  

\textbf{Proposition:  }  Let $\zeta^*$ and $\zeta^*_{DOI}$ be the optimal values of Eq \ref{LPVer}- Eq \ref{RMP_new_primal}, at termination of CG respectively. If $\Xi$ satisfies Eq \ref{novelDoiBound}, then $\zeta^*_{DOI} = \zeta^*$. 

\begin{proof}
Let $(\gamma^*, \xi^*)$ be an optimal solution to Eq \ref{RMP_new_primal}. If $\xi^*_{dz} = 0$ for all $d\in \mathcal{D}, z \in \mathcal{Z}_d$, then $\zeta^*_{DOI} = \zeta^*$ because $\gamma^*$ is feasible and optimal for Eq \ref{RMP_new_primal}. Otherwise, there exists an observation $d \in \mathcal{D}$, $g\in \mathcal{G}$ such that $\gamma^*_g > 0$, $G_{dg}=1$ and $(Z_{dzg}=1) \rightarrow (\xi^*_{dz}>0) \quad  \forall z \in \mathcal{Z}_d$. 
Let $z_*\leftarrow \max_{\substack{z \in \mathcal{Z}_d \\ Z_{dgz}=1}}z$.  
Let $\alpha = \min \> \{ \gamma^*_g, \xi^*_{dz_*} \}$. Consider the solution obtained from $(\gamma^*, \xi^*)$ by decreasing $\gamma^*_g$ and $\xi^*_{dz}$ for all $z \in \mathcal{Z}_d$ s.t. $(Z_{dgz}=1)$ by $\alpha$ and increasing $\gamma^*_{\bar{g}(g,\{d\})}$ by $\alpha$.  We have increased the objective by $\alpha (\Gamma_{\bar{g}(g,\{d\})} - \Gamma_g-\Xi_{dg})$ which is non-positive since  $\sum_{z \in \mathcal{Z}_d}\Xi_{dz}Z_{gzg}=\Xi_{dg}>  \Gamma_{\bar{g}(g,\{d\})} - \Gamma_g$
and $\alpha >0$. Thus $(\gamma^*, \xi^*)$ is feasible for Eq \ref{RMP_new_primal} and has a cost that is less than $\zeta^*_{DOI}$. This contradicts the optimality of $(\gamma^*, \xi^*)$ and proves that there is no $d \in \mathcal{D},z \in \mathcal{Z}_d$ such that $\xi^*_{dz} > 0$. 
\end{proof}
We can produce a feasible binary solution when $\gamma$ is fractional at termination of CG as follows. We solve Eq \ref{RMP_new_primal} over $\hat{\mathcal{G}}$, while enforcing $\gamma_g$ to be binary for all $g \in \hat{\mathcal{G}}$.  If the solution has active $\xi$ terms, then we apply the procedure described in the proof above to decrease the cost of the solution and ensure feasibility to Eq \ref{ILPVER}. 

As CG proceeds we can not consider all of $\mathcal{Z}_{d}$ since the cardinality $\mathcal{Z}_{d}$ may explode for some or all $d \in \mathcal{D}$.  Thus we use a subset of $\mathcal{Z}_{d}$ consisting of, the largest element and $K$  others selected uniformly across $\mathcal{Z}_{d}$ denoted $\hat{\mathcal{Z}}_d$ (where $K$ is a user defined parameter; e.g. $K=5$ works well).  Thus $\hat{\mathcal{Z}}_d= \{z_{\lceil \frac{k|\mathcal{Z}_d|}{K+1} \rceil}   \quad \forall 1 \leq k\leq K+1 \} $.  With some abuse of notation we have $Z_{dzg}$ be defined over $\Xi^{+}_{dg}$, where $\Xi^{+}_{dg}=   G_{dg}\min_{\substack{z \in \hat{\mathcal{Z}}_d \\ \omega_{dz} \geq \Xi_{dg} }}\omega_{dz}$.

%
\subsection{Efficient Pricing}
\label{novelReduce}
Pricing for Eq \ref{RMP_new_primal} is conducted as
$\min_{g \in \mathcal{G}}
    \Gamma_g- \sum_{\substack{d \in \mathcal{D} \\ z \in \mathcal{Z}_d }} Z_{dzg}\lambda_{dz}$.  
Current MWSP applications (as in \cite{mwspJournal}) are associated with mechanisms to solve Eq \ref{pricing_gen} instead of $\min_{g \in \mathcal{G}}
    \Gamma_g- \sum_{\substack{d \in \mathcal{D} \\ z \in \mathcal{Z}_d }} Z_{dzg}\lambda_{dz}$.  We now prove that doing pricing using Eq \ref{pricing_gen} where $\lambda_d\leftarrow \sum_{z \in \mathcal{Z}_d}\lambda_{dz}$ $\forall d \in \mathcal{D}$ ensures that Eq \ref{LPVer}=Eq \ref{RMP_new_primal} at termination of CG.

\textbf{Claim:  }
If $\lambda^*$ is a dual optimal solution to Eq \ref{RMP_new_dual} (defined over some $\hat{\mathcal{G}} \subseteq \mathcal{G}$) satisfying that Eq \ref{pricing_gen}$\geq 0$  then $ \sum_{\substack{d \in \mathcal{D} \\ z \in \mathcal{Z}_d }}\lambda^*_{dz}=\mbox{Eq } \ref{LPVer} $.

\textbf{Proof:  }
%
%
Since $\hat{\mathcal{G}} \subseteq \mathcal{G}$ then 
$\mbox{Eq } \ref{LPVer} \leq \sum_{ \substack{d \in \mathcal{D} \\z \in \mathcal{Z}_d }}\lambda^*_{dz}$.  Let $\lambda^+$ be defined as $\lambda^+_d=(\sum_{ z \in \mathcal{Z}_d }\lambda^*_{dz})$ $\forall d \in \mathcal{D}$. 
Since Eq \ref{pricing_gen}$\geq 0$ then $\lambda^+$ is a dual feasible solution to Eq \ref{RMP_DUAL} where $\mathcal{G}=\hat{\mathcal{G}}$ ; thus 
$\sum_{ d \in \mathcal{D} }\lambda^+_{d} \leq  \mbox{Eq } \ref{LPVer}$.  Since $\sum_{\substack{d \in \mathcal{D} \\ z \in \mathcal{Z}_d }}\lambda^*_{dz}=\sum_{ d \in \mathcal{D} }\lambda^+_{d}$ then we have lower and upper bounded $\sum_{\substack{d \in \mathcal{D} \\ z \in \mathcal{Z}_d }}\lambda^*_{dz}$ by Eq \ref{LPVer} establishing the claim.
\section{Application: Entity Resolution}
\label{SEC_Applications}
%
%
In this section we apply the MWSP formulation  in Section \ref{SEC_stab_mwsp} to entity resolution resulting in our approach, which we call F-MWSP.  This section is structured as follows. In Section \ref{pipeline_EC} we describe the problem domain of entity resolution, and outline our pipeline for solving such problems.  In Section \ref{App_clust_form} we define problem specific cost function for evaluating a single hypothesis in entity resolution.  In Section \ref{pricingAppSec} we devise efficient pricing algorithms (\ie\ finding hypotheses with negative reduced costs) that exploit structural properties of entity resolution.  In Section \ref{derivationNewXi} we describe the production of $\Xi_{dg}$ terms that satisfy Eq \ref{novelDoiBound}, thus defining the F-DOIs for entity resolution problems. In Section \ref{numeric} we provide a numerical example demonstrating the value of F-DOI in this application.  
%
\subsection{Pipeline for Entity Resolution}
\label{pipeline_EC}
%
Entity resolution seeks to construct a surjection from observations in input dataset to real world entities.  The observations in the dataset are denoted $\mathcal{D}$, as defined in Section~\ref{Sec_review_ilp}.  Specifically, the dataset consists of a structured table where each row (or tuple) represents an observation of a real world entity. We rely on the attributes of the table to determine if two observations represent the same real world entity.

A naive way of doing entity resolution is to compare every pair of observations in the input dataset and decide whether they belong to the same entity or not; this will result in $|\mathcal{D}| \choose 2 $ comparisons, which is often prohibitively large for real-world applications.  We instead employ a technique called blocking \citep{konda2016magellan}, in which we use a set of pre-defined, fast-to-run predicates to identify the subset of pairs of observations which could conceivably correspond to common entities (thus blocking operates in the high recall regime).

\begin{figure}[t]
\centering
\includegraphics[width=0.9\columnwidth]{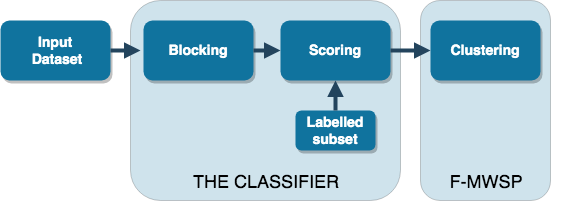} 
\caption{\textbf{Entity Resolution Pipeline} The stages of our pipeline are written in the following order.  Given our input dataset we apply blocking to produce a limited set of pairs of observations that may be co-associated in a common hypotheses.  Next we provide a probability score for each such pair using a classifier trained to distinguish between pairs that are/are not part of a common entity in the ground truth.  Finally we convert the output of the probability scores to cost terms and treat the entity resolution as a MWSP problem as described in Section \ref{SEC_stab_mwsp}.  
}
\label{fig:ERpipeline}
\end{figure}

We first use blocking to filter out majority of pairs of observations, which leaves only a small proportion of pairs for further processing.  Next, we generate a score for each pair of observations returned by the blocking step.  The probability score defined over a given pair of observations is the probability that the pair are associated with a common entity. The classifier that generates probability scores is trained by any learning algorithm on the annotated data. We take negative of probability scores and add a bias to them, forming the cost terms used in our MWSP algorithm.  Finally, based on the cost terms, the MWSP algorithm packs the observations into hypothesis with the goal of creating a bijection from hypothesis in the packing to real world entities. We refer to the combination of the blocker and the scorer as the \textit{classifier}.  Our entire pipeline for solving the entity resolution problem is described in Fig \ref{fig:ERpipeline}.
\subsection{Cost Function for Entity Resolution}
\label{App_clust_form}
Consider a set of observations $\mathcal{D}$, where for any $d_1 \in \mathcal{D},d_2 \in \mathcal{D}$ that $\theta_{d_1d_2} \in \mathbb{R}$ is the cost associating with including $d_1,d_2$ in a common hypothesis. Here positive/negative values of $\theta_{d_1d_2}$ discourage/encourage $d_1,d_2$ to be associated with a common hypothesis.  The magnitude of $\theta_{d_1d_2}$ describes the degree of discouragement/encouragement.  We assume without loss of generality $\theta_{d_1d_2}=\theta_{d_2d_1}$.  We construct $\theta_{d_1d_2}$ from the output classifier as $(0.5-p_{d_1d_2})$ where $p_{d_1d_2}$ is the probability provided by the classifier that $d_1,d_2$ are associated with a common hypothesis in the ground truth.  It is a structural property of our problem domain that most pairs of observations can not be part of a common hypothesis.  For such pairs $d_1,d_2$ then $\theta_{d_1d_2}=\infty$.  These are the pairs not identified by the blocker as being feasible.  We use $\theta_{dd}=0$ for all $d \in \mathcal{D}$.
We define the cost of a hypothesis $g \in \mathcal{G}$ as follows.  
\begin{align}
\label{eq_cost_entitty}
    \Gamma_g=\sum_{\substack{d_1 \in \mathcal{D} \\ d_2 \in \mathcal{D}}}\theta_{d_1d_2}G_{d_1g}G_{d_2g}
\end{align}
%
With the cost of a hypothesis defined, we can now treat entity resolution as a MWSP problem, and use CG to solve it.  Any observation not associated with any selected hypothesis in the solution to MWSP is defined to be in a hypothesis by itself of zero cost. Our formulation of entity resolution can also be rewritten as correlation clustering \citep{corclustorig}, which is usually tackled via LP relaxations with cycle inequalities and odd wheel inequalities \citep{nowozin2009solution} in the machine learning literature.  In the appendix we prove Eq \ref{LPVer} is no looser than \citep{nowozin2009solution}.

%
\subsection{Pricing}
\label{pricingAppSec}
With hypothesis cost $\Gamma_g$ defined in Eq~\ref{eq_cost_entitty}, we can now proceed to solve Eq~\ref{pricing_gen}.  However, solving Eq \ref{pricing_gen} would be exceedingly challenging if we had to consider all $d \in \mathcal{D}$ at once.  Fortunately, we can circumvent this difficulty using the following observation inspired by \citep{zhang2017efficient}, which studies biological cell instance segmentation.  For any fixed $d^* \in \mathcal{D}$, solving for the lowest reduced cost hypothesis that includes $d^*$ is much less challenging than solving Eq \ref{pricing_gen}.  This is because given $d^*$ all $d \in \mathcal{D}$ for which $\theta_{d^*d}=\infty$ can be removed from consideration.  Solving Eq \ref{pricing_gen} thus consists of solving many parallel pricing sub-problems, one for each $d^* \in \mathcal{D}$.  All negative reduced cost solutions are then added to $\hat{\mathcal{G}}$.  In this subsection we expand on this approach.

First we produce a small set of sub-problems each defined over a small subset of $\mathcal{D}$. Then we study \textit{exact} optimization of those sub-problems, followed by \textit{heuristic} optimization.
%
%
%
\subsubsection{Pricing Formulation of \citep{zhang2017efficient}}  
We write pricing sub-problem adapted from \citep{zhang2017efficient} given $d^* \in \mathcal{D}$ as follows:  
\begin{align}
\label{pricing_sub_local}
\min_{\substack{g \in \mathcal{G}\\ G_{dg}=0 \quad \forall d \notin \mathcal{D}_{d^*} \\ G_{d^*g}=1}}\Gamma_g-\sum_{d \in \mathcal{D}}\lambda_dG_{dg}\\
\mathcal{D}_{d^*}=\{ d \in \mathcal{D}; \theta_{dd^*}<\infty \} \nonumber
\end{align}
Here $\mathcal{D}_{d^*}$ is the set of observations that may be grouped with observation $d^*$, which we call its neighborhood.  Since the lowest reduced cost hypothesis must contain some $d^*\in \mathcal{D}$ by solving Eq \ref{pricing_sub_local} for each $d^*\in \mathcal{D}$ we solve Eq \ref{pricing_gen}.  

\subsubsection{Improving on \citep{zhang2017efficient} by decreasing sub-problem size} 
We improve on \citep{zhang2017efficient} by decreasing the number of observations considered in sub-problems, particularly those with large numbers of observations.  We achieve this by associating a unique rank $r_d$ to each observation $d \in \mathcal{D}$, such that $r_d$ increases with  $|\mathcal{D}_d|$, \ie\ the more neighbors an observation has, the higher rank it is assigned. To ensure that each observation has unique rank we break ties arbitrarily. 

Given that $d^*$ is the lowest ranking observation in the hypothesis we need only consider the set of observations s.t. $d \in  \{ \mathcal{D}_{d^*}\cap  \{r_d \geq r_{d^*}\} \}$, which we define to be $\mathcal{D}^*_{d^*}$.  
We write the resultant pricing sub-problem as follows.  
\begin{align}
\label{pricing_sub_local2}
\min_{\substack{g \in \mathcal{G}\\ G_{dg}=0 \quad \forall d \notin \mathcal{D}^*_{d^*}\\ G_{d^*g}=1}}\Gamma_g-\sum_{d \in \mathcal{D}}\lambda_dG_{dg}
\end{align}
%
%
\subsubsection{Further improving on \citep{zhang2017efficient} by  removing superflous sub-problems}
We can also decrease the number of sub-problems considered as follows.  First we relax the constraint $G_{d^*g}=1$ in Eq \ref{pricing_sub_local2}.  Now observe that for any $d_2\in \mathcal{D}$,$d \in \mathcal{D}$ s.t. $\mathcal{D}^*_{d} \subset \mathcal{D}^*_{d_2}$ that the lowest reduced cost hypothesis over $\mathcal{D}^*_{d_2}$ has no greater reduced cost than that over $\mathcal{D}^*_{d}$.  We refer a neighborhood $\mathcal{D}^*_{d^*}$ as being non-dominated if no $d_2 \in \mathcal{D}$ exists s.t. $\mathcal{D}^*_{d} \subset \mathcal{D}^*_{d_2}$ . 

During pricing we iterate over non-dominated neighborhoods.  For a given non-dominated neighborhood $\mathcal{D}^*_{d^*}$ we write the pricing sub-problem below.
%
\begin{align}
\label{pricing_sub_local3}
\min_{\substack{g \in \mathcal{G}\\ G_{dg}=0 \quad \forall d \notin \mathcal{D}^*_{d^*}}}\Gamma_g-\sum_{d \in \mathcal{D}}\lambda_dG_{dg}
\end{align}
\subsubsection{(A) Exact Pricing}
%
%
We now consider the exact solution of Eq \ref{pricing_sub_local3}.  We frame Eq \ref{pricing_sub_local3} as a ILP, which we solve using a mixed integer linear programming (MILP) solver.  
We use decision variables $x,y$ as follows. We set binary variable $x_d=1$ to indicate that $d$ is included in the hypothesis being generated and otherwise set $x_d=0$.  We set $y_{d_1d_2}=1$ to indicate that both $d_1,d_2$ are included in the hypothesis being generated and otherwise set $y_{d_1d_2}=0$. %
%
%
%
Defining $\mathcal{E}^- = \{ (d_1, d_2): \theta_{d_1 d_2} = \infty \}$ as the set containing pairs of observations that cannot be grouped together, and $\mathcal{E}^+ = \{ (d_1, d_2): \theta_{d_1 d_2} < \infty \}$ as the set containing pairs of observations that can be grouped together, we write the solution to Eq \ref{pricing_sub_local3} as a MILP, which we annotate below.    
%

\begin{align}
\label{initPricingObj}
   \min_{\substack{ x_d \in \{0,1\} \\ \forall d \in \mathcal{D}^*_{d^*}\\ y \geq 0}} & \sum_{d \in \mathcal{D}^*_{d^*}}-\lambda_d x_{d}
    +\sum_{\substack{d_1 \in \mathcal{D}^*_{d^*} \\ d_2 \in \mathcal{D}^*_{d^*} \\ (d_1, d_2) \in \mathcal{E}^+} }\theta_{d_1d_2}y_{d_1d_2}  \\
    \label{eq_pri_edge}
   \text{s.t.} \quad & x_{d_1}+x_{d_2} \leq 1  \quad \forall (d_1,d_2) \in \mathcal{E}^-\\
    \label{eq_pri_x_y_constist_neg1}
   & y_{d_1d_2}\leq x_{d_1} \quad \forall (d_1,d_2) \in \mathcal{E}^+\\
    \label{eq_pri_x_y_constist_neg2}
   & y_{d_1d_2}\leq x_{d_2} \quad  \forall (d_1,d_2) \in \mathcal{E}^+\\
    \label{eq_pri_x_y_constist_pos}
   & x_{d_1}+x_{d_2}-y_{d_1d_2}\leq 1 \quad \forall (d_1,d_2) \in \mathcal{E}^+
\end{align}

%
\textbf{Eq \ref{initPricingObj}}: Defines the reduced cost of the hypothesis being constructed.
\textbf{Eq \ref{eq_pri_edge}}:  Enforce that pairs for which $\theta_{d_1d_2}=\infty$ are not include in a common hypothesis.
\textbf{Eq \ref{eq_pri_x_y_constist_neg1}-Eq \ref{eq_pri_x_y_constist_pos}}: Enforce that $y_{d_1d_2}=x_{d_1}x_{d_2}$.  Observe that given that $x$ is binary, that $y$ must also be binary so as to obey Eq \ref{eq_pri_x_y_constist_neg1}-Eq \ref{eq_pri_x_y_constist_pos}. Thus we need not explicitly enforce $y$ to be binary.  
%
\subsubsection{(B) Heuristic Pricing}
Solving  Eq \ref{pricing_sub_local3} exactly using Eq \ref{initPricingObj}-Eq \ref{eq_pri_x_y_constist_pos} for each non-dominated neighborhood can be too time intensive for some scenarios. In fact Eq \ref{pricing_sub_local3} generalizes max-cut, which is NP-hard \cite{karp}.  This motivates the use of heuristic methods to solve Eq \ref{pricing_sub_local3}.  Heuristic pricing is commonly used in operations research, however \textit{we are the first paper in machine learning/entity resolution to employ this strategy}. Thus we decrease the computation time of pricing by decreasing the number of sub-problems solved, and solving those that are solved heuristically.  
\begin{itemize} 
\item \textbf{Early termination of pricing:  } Observe that solving pricing (exactly or heuristically) over a limited subset of the sub-problems produces an approximate minimizer of Eq \ref{pricing_gen}.  We decrease the number of sub-problems solved during a given iteration of CG as follows.  We terminate pricing in a given iteration when $M$  negative reduced cost hypothesis have been added to $\hat{\mathcal{G}}$ in that iteration of CG ($M$ is a user defined constant; $M=50$ in our experiments).  This strategy is called partial pricing \citep{lubbecke2005selected}
\item\textbf{Solving sub-problems approximately:  } We found empirical success solving Eq \ref{initPricingObj}-Eq \ref{eq_pri_x_y_constist_pos} using the quadratic pseudo-Boolean optimization with the improve option used (QPBO-I) \citep{QPBO}. 
\end{itemize}
  The use of heuristic pricing does not prohibit the exact solution of Eq \ref{LPVer}.  One can switch to exact pricing after heuristic pricing fails to find a negative reduced cost hypothesis in $\mathcal{G}$.
\subsection{Computing $\Xi_{dg}$ for Entity Resolution}
\label{derivationNewXi}
In this section, for any given $g \in \hat{\mathcal{G}}$ we construct  $\Xi_{dg}$ to satisfy Eq \ref{novelDoiBound}, which in practice leads to efficient optimization.  
We rewrite $\epsilon+\Gamma_{\bar g(g,\mathcal{D}_s)}-\Gamma_g$ by plugging in the expressions for $\Gamma_g$ in Eq \ref{eq_cost_entitty}.  We use $\mathcal{D}_{g}$ to denote the subset of $\mathcal{D}$ for which $G_{dg}=1$.  
\begin{align}
\label{eq_step_3_bopund}
    \epsilon 
    +\sum_{\substack{d_1 \in \mathcal{D}_g \\ d_2 \in \mathcal{D}_g}}-\theta_{d_1d_2}\max([d_1 \in \mathcal{D}_s],[d_2 \in \mathcal{D}_s])  
\end{align}
We now bound components of  Eq \ref{eq_step_3_bopund} as follows.  For $\theta_{d_1d_2}<0$ we upper bound $ -\theta_{d_1d_2}\max([d_1 \in \mathcal{D}_s],[d_2 \in \mathcal{D}_s])$ with:  
$-\theta_{d_1d_2}([d_1 \in \mathcal{D}_s]+[d_2 \in \mathcal{D}_s])$ 
%
For $\theta_{d_1d_2}>0$ we upper bound $ -\theta_{d_1d_2}\max([d_1 \in \mathcal{D}_s],[d_2 \in \mathcal{D}_s])$  with:
$ -\frac{\theta_{d_1d_2}}{2}([d_1 \in \mathcal{D}_s]+[d_2 \in \mathcal{D}_s])$
%
Below we plug the upper bounds into Eq \ref{eq_step_3_bopund}; group by $[d \in \mathcal{D}_s]$; and enforce non-negativity of the result.  $\mbox{Eq } \ref{eq_step_3_bopund} \leq \sum_{\substack{d \in \mathcal{D}}} [d \in \mathcal{D}_s]\Xi_{dg}$ where $\Xi_{dg}=0$ for $d \notin \mathcal{D}_g$, and is otherwise defined below. 
\begin{align}
     \Xi_{dg}= \epsilon+\max(0,-\sum_{\substack{d_1 \in \mathcal{D}_g}}\theta_{dd_1}(1+[\theta_{d d_1}<0]) ) \quad \forall d \in \mathcal{D}_g \nonumber 
\end{align}
The analysis in in this section can be applied directly for other pairwise cost functions such as multi-person pose estimation and multi-cell segmentation as described in \citep{mwspJournal}.
\subsection{Numerical Example}
\label{numeric}
We provide a motivating example demonstrating how the use of F-DOI provides a lower valued LP relaxation given fixed $\hat{\mathcal{G}}$ than using varying DOI or no DOI.  Consider a correlation clustering problem instance over $\mathcal{D}=\{d_1,d_2,d_3,d_4,d_5$ where $\theta$ is defined as follows.  
Let $\theta_{d_1,d_2}=\theta_{d_2,d_3}=\theta_{d_1,d_3}=\theta_{d_4d_5}=-100; \theta_{d_3,d_4}=\theta_{d_3d_5}=-1$ (note $\theta_{d \hat{d}}=\theta_{\hat{d}d}$). All other $\theta$ terms not defined above take on value $\infty$ .  Consider that $\hat{\mathcal{G}}=g_1,g_2$ where  $\mathcal{D}_{g_1}= \{d_1,d_2,d_3\}$ and $\mathcal{D}_{g_2}=\{d_3,d_4,d_5\}$. The optimal solution to the set packing RMP using no DOI is limited to selecting either $g_1$ or $g_2$. Solving the RMP using varying DOI both $g_1,g_2$ can be selected but a penalty of $400+\epsilon$ must be paid, resulting in higher objective than selecting only $g_1$. Solving the RMP using F-DOI both $g_1$, $g_2$ can be selected with a penalty of $4+\epsilon$, resulting in a lower objective than selecting either $g_1$ or $g_2$. 

%
%

%
\section{Experiments}
\label{SEC_Exper}
In this section, we study the different properties of the F-MWSP clustering algorithm and evaluate the performance scores on certain benchmark datasets. The classifier, which encompasses the blocker and the scorer, is a crucial component of the entity resolution pipeline (see Figure~\ref{fig:ERpipeline} and Section~\ref{pipeline_EC}). We leverage the methods provided in a popular and open source entity resolution library called Dedupe \citep{dedupe} to handle the blocking and scoring functionalities for us. Dedupe offers attribute type specific blocking rules and a ridge logistic regression algorithm as a default for scoring. Certainly, a more powerful classifier, especially if designed keeping the domain of the dataset in mind, can significantly boost the performance of the clustering outcome. As the focus of this paper has been F-MWSP clustering algorithm, an intuitive and reasonably good classifier such as Dedupe suits our setting.\\
In the following sections, we first demonstrate the different properties of F-MWSP algorithm on a single dataset and then compare its performance with other methods on benchmark datasets.

\subsection{Characteristics of F-MWSP algorithm}
\label{sec:exp_char}
\subsubsection{The Setting}
\label{sec:exp_char_setting}
To understand the benefits of F-MWSP clustering, it will be helpful to first conduct ablation study on a single dataset. The dataset that we choose in this section is called \textit{patent\_example} and is publicly available on Dedupe.  \textit{patent\_example} is a labelled dataset listing the patent statistics of the Dutch innovators. It has has $2379$ entities and $102$ clusters where the mean size of the cluster is $23$. We split the dataset into two halves and set aside the second half only to report the accuracies. The first half of the dataset that is visible to the learning algorithm from which we randomly sample about $1\%$ of the total matches and provide it to the classifier as a labelled data. 
\subsubsection{ (A) Superior performance over hierarchical clustering}
Table~\ref{tab: patstat_baseline} shows that F-MWSP clusters offers better performance over hierarchical clustering, a standard method of choice for clustering problems \citep{hastie2005elements}.  The performance has been evaluated against standard clustering metrics.

\begin{table}[!t]
	\centering
	\resizebox{0.75\columnwidth}{!}{%
	\begin{tabular}{c c c} 
	 Performance & Hierarchical & F-MWSP \\  
	 Metric & Clustering & Clustering \\ 
	 \hline\hline
	  Precision / Recall &  95.5\% / 89.1\% & 95.4\% / 94.3\% \\
	  F1 measure & 92.2\% & 94.8\% \\ \hline
	  Homogn. / Compltn. &  94.6\% / 94\% & 94.4\% / 96.5\% \\
	  V measure & 94.3\% & 95.4\% \\ \hline
	  Adjusted Rand Index & 91.3\% & 94.2\% \\ \hline
	  Fowlkes Mallows & 92.2\% & 94.8\% \\
	 \hline\hline
	\end{tabular}
	}
	\label{tab: patstat_baseline}
	\caption{F-MWSP algorithm performs better than the baseline hierarchical clustering algorithm.}
\end{table}

\subsubsection{ (B) Significant speed-ups owing to Flexible DOIs} 
\begin{figure}[!t]
\centering
\includegraphics[width=0.75\columnwidth]{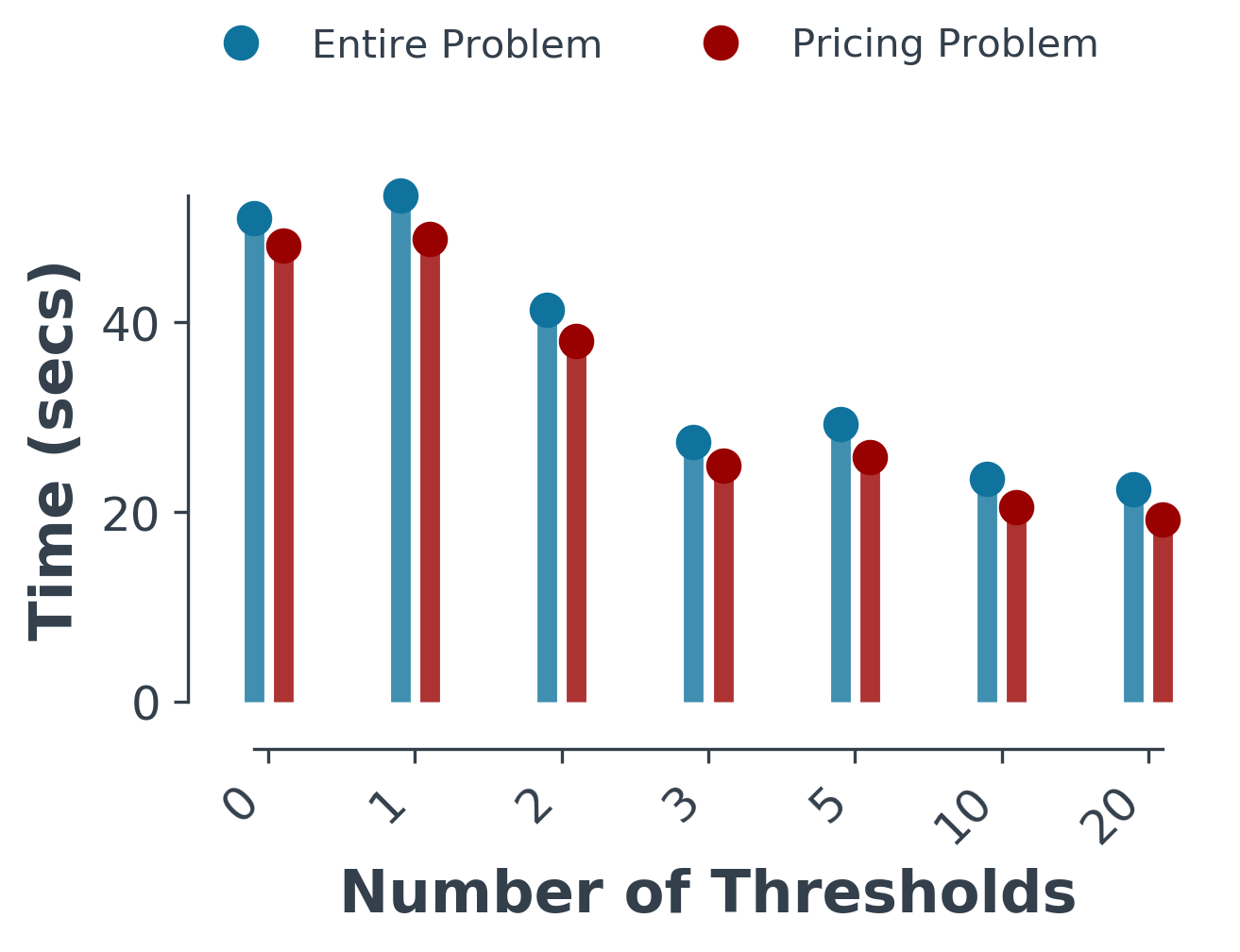} 
\caption{\textbf{Speedups with Flexible DOIs} Varying the number thresholds (value of $K$) of the Flexible DOIs improves the convergence speed. Threshold value $0$ corresponds to the Varying DOIs used in \citep{mwspJournal}. }
\label{fig:speedups}
\end{figure}

We obtain at least $20\%$ speed up with our proposed Flexible DOIs over Varying DOIs \citep{mwspJournal} as indicated in Figure~\ref{fig:speedups}. Moreover, we also observe that the computation time of the problem decreases as the number of thresholds (value of K) increases, with up to $60\%$ speedup.   

\subsubsection{ (C) Tractable solutions to the pricing problem}
Recall the strategies discussed to solve the pricing problem from Section~\ref{pricingAppSec}, namely, exact and heuristic. Exact pricing is often not feasible in entity resolution owing to the large neighborhoods of some sub-problems. Fortunately, the heuristic solver helps cut down the computation time by a large fraction. For instance, \textit{patent\_example} experiment takes atleast \textbf{1 hour} for completion with the exact solver while with the heuristic solver it takes about \textbf{20 seconds}.

\subsection{F-MWSP algorithm on benchmark datasets}
To make sure that our findings are broadly applicable, we conducted experiments with more entity resolution benchmark datasets. \citep{saeedi2017comparative} provides us with some interesting entity resolution datasets which we also include in this section. The statistics of all the datasets used in the paper are available in Table~\ref{tab:datasetstats}.
\subsubsection{The Setting}
We make our setting consistent with \citep{saeedi2017comparative} to be able to compare against their clustering algorithms. \citep{saeedi2017comparative} leverages hand-crafted rules designed on the entire dataset to generate the cost terms. The costs are then fed into various clustering algorithms and the performance is evaluated over the whole dataset. We use dedupe classifier which is trained on a small percentage of matches from a split half of the dataset similar to Section~\ref{sec:exp_char_setting}. F-MWSP is then evaluated based on the dedupe cost terms over the entire dataset. 

\begin{table}[!t]
	\centering
	\resizebox{\columnwidth}{!}{%
		\begin{tabular}{c c c c c c} 
			Dataset & Entities & Matches & Clusters & Mean & Max\\ [0.5ex]
			\hline\hline
			patent\_example & 2379 & 293785 & 102 & 23 & 676 \\ \hline
			csv\_example & 3337 & 6608 & 1162 & 3 & 18 \\ \hline
			Affiliations & 2260 & 16795 & 330 & 7 & 47 \\ \hline
			Settlements & 3054 & 4388 & 820 & 3.7 & 4  \\ \hline
			Music 20K & 19375 & 16250 & 10000 & 2 & 5 \\ \hline\hline
		\end{tabular}
	}
	\caption{{\bf Dataset statistics} The statistics of all the datasets used in the paper are presented here. \textit{Mean} and \textit{Max} denote the respective statistics over the cluster sizes.}
	\label{tab:datasetstats}
\end{table}

 \begin{table}[!t]
	\centering
	\resizebox{0.5\columnwidth}{!}{%
		\begin{tabular}{c c c c c} 
			Method  & Settlements & Music 20K  \\ [0.5ex]
			\hline\hline
			ConCom  & 0.65 & 0.26  \\ \hline
			CCPivot  & 0.90 & 0.74  \\ \hline
			Center  & 0.88 & 0.66   \\ \hline
			MergCenter  & 0.68 & 0.39  \\ \hline
			Star1  & 0.82 & 0.62  \\ \hline
			Star2  & 0.92 & 0.69 \\ \hline
			F-MWSP  & \textbf{0.96} & \textbf{0.81} \\ \hline\hline
		\end{tabular}
	}
	\caption{\textbf{F-MWSP on benchmark datasets} We obtain higher F1 score over the methods reported in \citep{saeedi2017comparative}. The F1 scores for other methods are extracted from the paper's bar plot.}
	\label{tab:mwsp_sota}
\end{table}
\subsubsection{F-MWSP is competitive}
In this section, we report the performance of F-MWSP clustering on different datasets and compare against the baselines available for them. We start with \textit{csv\_example} dataset which is publically available on Dedupe akin to \textit{patent\_example}. On \textit{csv\_example}, F-MWSP achieves a higher F1 score of \textbf{95.2 \%} against hierarchical clustering 94.4\%, the default in Dedupe. In Table~\ref{tab:mwsp_sota}, we compare the performance of our entity resolution pipeline against algorithms in \citep{saeedi2017comparative}. Table~\ref{tab:mwsp_sota}  demonstrates that our pipeline, with F-MWSP clustering, is as powerful as the recognized entity resolution algorithms.
The \textit{affiliations} dataset used in \citep{aumueller2009web} which is unique in the sense that the lack of structure in the data generates poorer cost terms. Despite this, F-MWSP gives us an F1 score of \textbf{$63\%$}, however, we note that a well handcrafted rule-based classifier improve the F1 score as demonstrated in  \citep{aumueller2009web}.
\section{Conclusions}
\label{SEC_conc}
In this paper we formulate entity resolution as MWSP problem.  To solve such a problem, we devise a novel CG formulation that employs flexible dual optimal inequalities which use hypothesis specific information when constructing dual bounds.  Our formulation exploits the fact that most pairs of observations can not be in a common hypothesis to produce pricing subproblems over small subsets of the observations that can be easily solved for some datasets, and for others can be solved to high quality heuristically.  
We demonstrate superior performance to the baseline hierarchical clustering formulation to entity resolution.

\subsubsection{Acknowledgement}
We thank Dr. Vikas Singh from UW Madison and Dr. Konrad Kording from UPenn for the discussions on the project.
Shaofei Wang is supported by NIH grant R01NS063399. Julian Yarkony acknowledges the help provided by Verisk.

\section{Appendix}
\label{MCRelationship}
%
In this section we describe Correlation Clustering (CC) \citep{corclustorig}, its standard relaxation and finally prove that our MWSP relaxation of CC is tighter than the standard relaxations in the literature \citep{nowozin2009solution,ilpalg,highcc}. 

Given a graph with node set $\mathcal{D}$ we use $f \in \{0,1\}^{|\mathcal{D}| \times |\mathcal{D}|}$ which we index by $d_1,d_2$ to describe a partition of $\mathcal{D}$.  We set $f_{d_1d_2}=1$ if and only if $d_1,d_2$ are in a common component in our solution. We use $\theta_{d_1d_2}$ to denote the cost of including $d_1,d_2$ in a common component.  The objective of CC is written below.
\begin{align}
\label{objCC}
        \min_{f \in \{0,1\}}\sum_{\substack{d_1 \in \mathcal{D} \\ d_2 \in \mathcal{D}}} \theta_{d_1d_2}f_{d_1d_2}
\end{align}
CC uses cycle inequalities to  enforce that $f$ describes a valid partitioning of the vertices.  Cycle inequalities state that for every cycle of nodes, that the number of ``cut" edges (meaning edges where the connected observations are in separate components) is a number other than one.  
Let $\mathcal{H}$ be the set of cycles of vertices, which we index by $h$.  We treat $h$ as the set of edges on the associated cycle.  The cycle inequality associated with any $h \in \mathcal{H}$, $f_{d_ad_b} \in h$ is written below.   
\begin{align}
\label{cycIneqDef}
        \sum_{(d_1,d_2) \in h -(d_a,d_b)} (1-f_{d_1d_2}) \geq (1-f_{d_ad_b})
\end{align}
A solution $f$ must satisfy Eq \ref{cycIneqDef} for all $h \in \mathcal{H}$, $(d_a,d_b) \in h$ to be a feasible partition of the observations \citep{nowozin2009solution}.  
%
%
%
It is sufficient to enforce only the cycle inequalities over each cycle of three nodes in order to enforce all cycle inequalities \citep{chopra1993partition}.  We write the cycle inequality over observations $d_1,d_2,d_3$ below.  
\begin{align}
\label{cycIneqCC}
(1-f_{d_1d_3})+(1-f_{d_2d_3}) \geq (1-f_{d_1d_2})
\end{align}
%
%
Odd wheel inequalities (OWI) are a common valid inequality in CC used to tighten the LP relaxation of CC.  
OWI are defined over a cycle of edges of odd cardinality $h_b$ with a single additional node $d_b$ connected to all other nodes in the center.  We define OWI below for any $b \in \mathcal{B}$ where $\mathcal{B}$ is the set of OWI. We use $d^b_m$ to denote the $m$'th observation in the cycle $h_b$; and note that $d^b_{|h_b|+1}=d^b_1$.  
\begin{align}
\label{oddWheelDef}
\sum_{m=1}^{|h_b|} f_{ d^b_md_b}-f_{d^b_{m}d^b_{m+1}} \leq \lfloor \frac{|h_b|}{2} \rfloor
\end{align}
We produce an LP relaxation of CC by relaxing $f$ to lie in $[0,1]$ and enforcing Eq \ref{cycIneqCC}- Eq \ref{oddWheelDef}.
\begin{align}
\label{CCVERLPWODD}
\min_{ 1\geq f\geq 0}\sum_{\substack{d_1 \in \mathcal{D} \\ d_2 \in \mathcal{D}}} \theta_{d_1d_2}f_{d_1d_2}\\
(1-f_{d_1d_3})+(1-f_{d_2d_3}) &\geq (1-f_{d_1d_2}) \nonumber \\
     \sum_{m=1}^{|h_b|} f_{d^b_md_b}-f_{d^b_{m}d^b_{m+1}} &\leq \lfloor \frac{|h_b|}{2} \rfloor \nonumber 
\end{align}
For any $d_1\in \mathcal{D},d_2\in \mathcal{D}$ the variable $f_{d_1d_2}$ corresponds to the following in our MWSP formulation for EC.
\begin{align}
\label{fdMapping}
    f_{d_1d_2}=\sum_{g \in \mathcal{G}}G_{d_1g}G_{d_2g}\gamma_{g} 
\end{align}
The goal of the remainder of this section is to show that Eq \ref{LPVer} is no looser a relaxation of CC than Eq \ref{CCVERLPWODD}.  We outline this section as follows. In Section \ref{BoundsObeyed} we show that any  feasible solution to Eq \ref{LPVer} satisfies $0\leq f\leq 1$.  In Section \ref{APP_proof_mc} we show that any  feasible solution to Eq \ref{LPVer} satisfies Eq \ref{cycIneqCC}.  In Section \ref{APP_proof_odd} that that any  feasible solution to Eq \ref{LPVer} satisfies Eq \ref{oddWheelDef}.  In Section \ref{finalTIght} we establish that Eq \ref{LPVer} $\geq $ Eq \ref{CCVERLPWODD}.
\subsection{Proof Bound Obeyed}
\label{BoundsObeyed}
Eq \ref{CCVERLPWODD} enforces that for each $1\geq f_{d_1d_2}\geq 0$ and we now establish that this holds for Eq \ref{LPVer}.  Using Eq \ref{fdMapping} we observe that the following must hold for any $\gamma$ satisfying Eq \ref{LPVer}.  
\begin{align}
\label{lbPor}
    0 \leq \sum_{g \in \mathcal{G}}G_{d_1g}G_{d_2g}\gamma_{g}=f_{d_1d_2}\\
\label{ubPor}
    \sum_{g \in \mathcal{G}}G_{d_1g}G_{d_2g}\gamma_{g}=f_{d_1d_2} \leq 1 
\end{align}
 Eq \ref{lbPor} is satisfied since $G$ is a binary matrix and $\gamma$ is non-negative.  Eq \ref{ubPor} is satisfied  since $\sum_{g \in \mathcal{G}}G_{d_1g}G_{d_2g}\gamma_{g}\leq \sum_{g \in \mathcal{G}}G_{d_1g}\gamma_{g}$ and Eq \ref{LPVer} ensures that $\sum_{g \in \mathcal{G}}G_{d_1g}\gamma_{g}\leq 1$ for each $d_1 \in \mathcal{D}$.
\subsection{MWSP Satisfies All Cycle Inequalities }
\label{APP_proof_mc}
%
In this section we establish that any feasible solution $\gamma$ to Eq \ref{LPVer} satisfies Eq \ref{cycIneqCC}.  To assist in our discussion we use the notation $j^{\mathcal{D}^-}_{\mathcal{D}^+}$ to denote the sum of the $\gamma$ terms associated with hypothesis that include all elements in $\mathcal{D}^+$ and no elements in $\mathcal{D}^-$ as follows. 
\begin{align}
\label{Eq_z_def}
    j^{\mathcal{D}^-}_{\mathcal{D}^+}=\sum_{g \in \mathcal{G}}\gamma_g (\prod_{d \in \mathcal{D}^+}G_{dg})(\prod_{d \in \mathcal{D}^-}(1-G_{dg}))
\end{align}
We now use proof by contradiction to establish that $\gamma$ obeys Eq \ref{cycIneqCC}.  

\textbf{Claim: } 

All $\gamma$ satisfying Eq \ref{LPVer} satisfy all  inequalities of the form in Eq \ref{cycIneqCC}.

\textbf{Proof:}   
Suppose the claim is false.  Thus there exists a $\gamma$ that is feasible to Eq \ref{LPVer} for which there exists a $d_1,d_2,d_3$  that does not satisfy Eq \ref{cycIneqCC}.  We re-write Eq \ref{cycIneqCC} for the violated cycle inequality using $j$.
\begin{align}
\label{Eq_tripCycl}
    (1-j_{d_1d_3})+(1-j_{d_2d_3})< (1-j_{d_1d_2})\\
    1+j_{d_1d_2}< j_{d_1d_3}+j_{d_2d_3}\nonumber  \\
    1+j_{d_1d_2d_3}+j^{d_3}_{d_1d_2}<j_{d_1 d_3}^{d_2}+j_{d_2d_3}^{d_1}+2j_{d_1 d_2 d_3}\nonumber  \\
    1+j^{d_3}_{d_1d_2}<j_{d_1 d_3}^{d_2}+j_{d_2d_3}^{d_1}+j_{d_1 d_2 d_3} \nonumber 
    \end{align}
    We now bound the RHS by $j_{d3}$, which we in turn bound by 1.  
    \begin{align}
    1+j^{d_3}_{d_1d_2}<j_{d_1 d_3}^{d_2}+j_{d_2d_3}^{d_1}+j_{d_1 d_2 d_3}\leq j_{d_3}\leq 1 
\end{align}
Since $j^{d_3}_{d_1d_2}$ is non negative it can not be less than zero thus establishing a contradiction.
\subsection{MWSP Satisfies All Odd Wheel Inequalities}
\label{APP_proof_odd}
We now establish that all OWI are satisfied for any feasible solution to Eq \ref{LPVer} using proof by contradiction. 

    \textbf{Claim}
    \begin{align}
    \label{claimOdd}
     \sum_{m=1}^{|h_b|} j_{d^b_md_b}-j_{d^b_md^b_{m+1}} \leq \lfloor \frac{|h_b|}{2} \rfloor   \quad \forall b \in \mathcal{B}
    \end{align}
    \textbf{Proof: } 
    Consider a solution $\gamma$ and $b \in \mathcal{B}$ violating the claim. %
    \begin{align}
      \sum_{m=1}^{|h_b|} j_{d^b_md_b}-j_{d^b_md^b_{m+1}} > \lfloor \frac{|h_b|}{2} \rfloor  \nonumber \\
      \sum_{m=1}^{|h_b|} j_{d^b_{m}d_bd^b_{m+1}}+j^{d^b_{m+1}}_{d^b_{m}d_b} -j^{d_b}_{d^b_{m}d^b_{m+1}}-j_{d^b_md^b_{m+1}d_b} >  \lfloor \frac{|h_b|}{2} \rfloor \nonumber  \\
      \label{lastline}
      \sum_{m=1}^{|h_b|}  j^{d^b_{m+1}}_{d^b_{m}d_b}-j^{d_b}_{d^b_{m}d^b_{m+1}} > \lfloor \frac{|h_b|}{2} \rfloor    
    \end{align}
    We upper bound the LHS of Eq \ref{lastline} by removing the  $j^{d_b}_{d^b_{m}d^b_{m+1}}$ terms. $   \sum_{m=1}^{|h_b|}  j^{d^b_{m+1}}_{d^b_{m}d_b} >  \lfloor \frac{|h_b|}{2} \rfloor   $.  
We express $j$ using Eq \ref{Eq_z_def}.  
    \begin{align}
              \sum_{m=1}^{|h_b|} \sum_{g \in \mathcal{G}}\gamma_gG_{d^b_mg}(1-G_{d^b_{m+1}g})G_{d_bg}>  \lfloor \frac{|h_b|}{2} \rfloor   \nonumber \\
              \sum_{g \in \mathcal{G}}\gamma_g G_{d_bg}\sum_{m=1}^{|h_b|}G_{d^b_{m}g}(1-G_{d^b_{m+1}g})>  \lfloor \frac{|h_b|}{2} \rfloor   \nonumber 
    \end{align}
    Observe that the term $\sum_{m=1}^{|h_b|}G_{d^b_{m}g}(1-G_{d^b_{m+1}g})$ is bounded from above by $ \lfloor \frac{|h_b|}{2} \rfloor$.  This is because the largest independent set defined on a cycle graph contains half the nodes (rounded down).  We apply this bound below.
\begin{align}
\sum_{g \in \mathcal{G}}\gamma_gG_{d_bg}\lfloor \frac{|h_b|}{2} \rfloor> \lfloor \frac{|h_b|}{2} \rfloor \nonumber \\
\label{finEqOddProof}
\sum_{g \in \mathcal{G}}\gamma_gG_{d_bg}> 1 
\end{align}    
Eq \ref{LPVer} ensures that $\sum_{g \in \mathcal{G}}G_{dg}\gamma_g  \leq 1$  for all $d \in \mathcal{D}$ which  contradicts  Eq \ref{finEqOddProof} thus proving that the claim in Eq \ref{claimOdd} true. 
\subsection{Eq \ref{LPVer} $\geq $ Eq \ref{CCVERLPWODD} }
\label{finalTIght}
Since every feasible solution to Eq \ref{LPVer} obeys all constraints in Eq \ref{CCVERLPWODD} then the minimal cost solution to Eq \ref{LPVer} obeys all constraints in Eq \ref{CCVERLPWODD} thus Eq \ref{LPVer} $\geq $ Eq \ref{CCVERLPWODD}.  We have not established the existence of cases for which Eq \ref{LPVer} $>$ Eq \ref{CCVERLPWODD} and leave consideration of such cases to future research.  

\bibliographystyle{aaai} 
\bibliography{ms} 
 
\end{document}